\documentclass[11pt]{article}
\usepackage{fullpage, amsmath, amsfonts, amsthm, amssymb, natbib, tikz, bbm}
\usepackage{graphicx, enumerate, color, hyperref}
\usepackage{algorithm, algpseudocode}
\usepackage{mathtools}
\usepackage{multirow}
\usepackage{float}
\usepackage{array}
\usepackage{caption}
\usepackage{hyperref}
\usepackage[flushleft]{threeparttable}
\usepackage{setspace}
\usepackage{subfig}
\usepackage{tabularx}
\usepackage{bm}
\usepackage{multirow}
\usepackage{algorithm, algpseudocode}
\usepackage{hyperref}
\usepackage{booktabs}
\algnewcommand\algorithmicinput{\textbf{Input:}}
\algnewcommand\Input{\item[\algorithmicinput]}
\algnewcommand\algorithmicoutput{\textbf{Output:}}
\algnewcommand\Output{\item[\algorithmicoutput]}
\algnewcommand{\algorithmicand}{\textbf{ and }}
\algnewcommand{\algorithmicor}{\textbf{ or }}
\algnewcommand{\OR}{\algorithmicor}
\algnewcommand{\AND}{\algorithmicand}
\newcommand{\StatexIndent}[1][3]{
  \setlength\@tempdima{\algorithmicindent}
  \Statex\hskip\dimexpr#1\@tempdima\relax}
\algdef{S}[IF]{IfNoThen}[1]{\algorithmicif\ #1}

\allowdisplaybreaks % allow breaking lines for equations running across everal lines

\newcommand{\vect}[1]{\mathbf{\bm{#1}}}

\newcommand{\removed}[1]{}
\newcommand{\activatebar}{%
  \begingroup\lccode`\~=`\|
  \lowercase{\endgroup\let~}\innermid 
  \mathcode`|=\string"8000
}
\newsavebox{\mybox}\newsavebox{\mysim}
\newcommand{\distras}[1]{%
  \savebox{\mybox}{\hbox{\kern3pt$\scriptstyle#1$\kern3pt}}%
  \savebox{\mysim}{\hbox{$\sim$}}%
  \mathbin{\overset{#1}{\kern\z\resizebox{\wd\mybox}{\ht\mysim}{$\sim$}}}%
}
\title{On a Bernoulli Autoregression Framework\\ for Link Discovery and Prediction}
\author{Xiaohan Yan\thanks{Both authors contributed equally to this research.} \qquad\quad Avleen S. Bijral\footnotemark[1]\\Microsoft, One Microsoft Way, Redmond, WA 98052\\ {\tt \{xiaoya,avbijral\}@microsoft.com}}
\date{\today}

\begin{document}
\maketitle
%% Abstract
\begin{abstract}
We present a dynamic prediction framework for binary sequences that is based on a Bernoulli generalization of the auto-regressive process. Our approach lends itself easily to variants of the standard link prediction problem for a sequence of time dependent networks. Focusing on this dynamic network link prediction/recommendation task, we propose a novel problem that exploits additional information via a much larger sequence of auxiliary networks and has important real-world relevance. To allow discovery of links that do not exist in the available data, our model estimation framework introduces a regularization term that presents a trade-off between the conventional link prediction and this discovery task. In contrast to existing work our stochastic gradient based estimation approach is highly efficient and can scale to networks with millions of nodes. We show extensive empirical results on both actual product-usage based time dependent networks and also present results on a Reddit based data set of time dependent sentiment sequences.
\end{abstract}

%%%%%%%%%%%%%%%%%%%%%%%%%%%%%%%%%
%                                         Introduction                                            %
%%%%%%%%%%%%%%%%%%%%%%%%%%%%%%%%%
\section{Introduction}\label{intro}
Predicting and recommending links in a future network is a problem of significant interest in many communities with problems of interest ranging from recommending friends \cite{aiello2012friendship}, patent partners \cite{wu2013patent} to applications in disease propagation \cite{myers2010convexity} and detection of abnormal communication \cite{huang2009time}. From a higher perspective link prediction is not markedly different from the problem of matrix completion when the network is represented as a adjacency matrix. However, in many cases past link information is predictive of the future. In one such application, \citet{richard2010} proposed an approach for time dependent graphs where past adjacency matrices and a forecast of a network feature sequence is used to estimate links at a future time. This is a dynamic extension of the static link prediction problem \cite{liben2007link} with origins in using network properties of a static snapshot to predict links. As a simple example consider the problem of predicting communication between two users over a network, past information is clearly predictive.

In this paper we discuss one variant of the dynamic link prediction problem and in contrast to \cite{richard2010} present a highly scalable model with an efficient estimation approach. The underlying idea behind our approach is to exploit a sequence of auxiliary networks with feature data to inform links in the network sequence of interest. This is a fairly common scenario in a large corporation with multiple products and services, it can be immensely useful to use network information gathered from a product A to inform usage (via link prediction) on a product B. In the online retail setting\footnote{Example dataset: \url{https://www.kaggle.com/retailrocket/ecommerce-dataset}} such as Amazon a variety of user-item interactions are possible including ``\textit{view}'', ``\textit{add-to-cart}'' and ``\textit{purchase}''. One could generate networks at a user-item level for each type of event, and leverage the ``\textit{view}'' network or the ``\textit{add-to-cart}'' network to infer connections on the ``\textit{purchase}'' network, i.e., when a user eventually buys an item. In such scenarios we have two network sequences, a main sequence that is expected to be sparse and a relatively dense auxiliary sequence with additional feature information that is predictive of a link in the main network. The introduction of the auxiliary sequences serves two purposes, one to provide candidates of links to predict from and a feature set that is possibly predictive of these links. However, the model we propose is general and holds even for a single network sequence with feature information. In the standard matrix completion framework, the observed entries are directly modeled with low rank constraints on the matrix. Instead, we model link probabilities using a Bernoulli parametrization, incorporate an auto-regressive structure to allow for dependence on past data and also include an additional dependence on logit transformed feature data. The equivalence of a low rank constraint in this framework is a relatively under-parametrized model as we shall see in Section \ref{sec:Model}.

%%%%%%%%%%%%%%%%%%%%%%%%%%%%%%%%%
%                                         Related Work                                           %
%%%%%%%%%%%%%%%%%%%%%%%%%%%%%%%%%
\section{Related Work}\label{sec:literature}
Existing work on link prediction has primarily focused on static networks with methods ranging from matrix factorization \cite{koren2008factorization}, node neighborhood measures \cite{sarkar2011theoretical,liben2007link} to network diffusion \cite{myers2010convexity}. For an extensive survey see \cite{wang2015link}. In contrast (dynamic) link prediction in the presence of a sequence of networks is relatively unexplored. \citet{vu2011continuous} propose a longitudinal network evolution approach that models edges at time $t$ as a multivariate counting process where the intensity function incorporates edge level features (network derived) using either the multiplicative Cox or additive Aalen form. Our setting is different in that we have two sequences of networks that we can exploit and the design of our estimation method enables us to discover new links which is not possible here. Moreover, the estimation methods require computation of large 
%$\mathcal{O}\left(|E_t|d\right)$ 
weight matrices and it is not clear if this can scale to networks with millions of nodes. Additionally, many real world networks tend to have low rank structures due to similar local node characteristics and explicit introduction of these constraints is a desirable feature. Closer in structure to our model is the nonparametric approach described in \cite{sarkar2012nonparametric}, the edge probabilities are assumed to be Bernoulli distributed with a link function connecting the edge/node probabilities to the node/edge level features. Besides the difference in the problem setting, \citet{sarkar2012nonparametric} use a nonparametric estimator for the link function that uses a kernel weighted local neighborhood approach (in time and feature space) to estimate the link function. This requires a search over edge features that are similar to the edge being modeled and can be quite expensive, even with the use of fast search techniques it is unlikely to scale to extremely large graphs. 

In the matrix completion line of work \citet{richard2010} propose a graph feature tracking (GFT) framework for dynamic networks where the entire network at a future step is estimated by solving an appropriately regularized optimization problem. The optimization reflects a trade-off between minimizing the error between the estimated and the last adjacency matrix, a low rank constraint and a term involving the features. This term attempts to minimize the gap between the mapped features for the estimated matrix and the unknown adjacency matrix at the future time. The main intuition being that the evolving features are indicative of links in the sequence of networks. This combination of optimization terms allows the method to uncover links that are not previously observed. See Section \ref{sec:simu} for more details of GFT considered in the simulation study. In this work we also introduce a feature regularization term to allow for link discovery but unlike \cite{richard2010} our approach is highly scalable. While their method involves an expensive SVD operation at every iteration we propose a stochastic gradient based algorithm with each iteration involving simple operations. A later work \cite{richard2014link} extended the GFT framework to allow for features that evolve as a vector auto-regressive process and as such is not computationally feasible for very large networks.

Perhaps most importantly our problem setup is more general and applies to \emph{arbitrary} binary sequences with mapped features. In Section \ref{sec:reddit} we apply our approach on a hyperlink network from Reddit and show its improved performance over a baseline approach. The Reddit hyperlink network represents a single network sequence and the applicability of our approach demonstrates its potential wide usage on modeling time-dependent binary sequences. The rest of the work is organized as follows. In Section \ref{sec:Model} we propose the BAR model with its estimation problem in Section \ref{sec:est}. We compare the BAR model with baselines (GFT, logistic) in a simulation study in Section \ref{sec:simu}. On two real datasets in Section \ref{sec:exp}, one from actual products \textit{A} and  \textit{B} \footnote{To maintain anonymity we don't reveal the actual product names.} and the other from Reddit, we apply the proposed model to show its superior performance in link prediction and discovery. Finally, we conclude the work with discussions in Section \ref{sec:conclude}.

%%%%%%%%%%%%%%%%%%%%%%%%%%%%%%%%%
%                                           BAR Model                                            %
%%%%%%%%%%%%%%%%%%%%%%%%%%%%%%%%%
\section{The BAR Model}\label{sec:Model}
Let the adjacency matrices $\vect{A}^{(t)} \in \{0,1\}^n$ and $\vect{B}^{(t)} \in \{0,1\}^N$ be the sequence of main and auxiliary networks for $t \in \{1,\dots,T\}$. Corresponding to the edges in the auxiliary sequence we also have a sequence of feature tensors $\vect{F}^{(t)}\in \Re^{N\times N\times d}$ such that, for all $(i,j) \in \vect{B}^{(t)}$ we have $\vect{F}^{(t)}_{ij} \in \Re^d$. We assume that an edge in a main network sequence exists at least once in the auxiliary sequence and that the number of nodes ($N$ and $n$) is fixed over time. This is not an issue in practice since we can always set $N$ to be the largest values over $t$ and assume all $0$ rows at other times.

There are two important aspects of our model definition. First, we believe that the feature sequence and the history of connections are predictive of a link in the future. As an example, in the context of an email network this could imply that an increased exchange between two nodes is possibly indicative of a future calendar event. Second, the probability of link formation evolves with time and in that sense depends on the entire sequence of edge level features. For any link $\vect{A}^{(t)}_{ij}$ we have
\begin{equation}\label{eq:BAR}
\begin{gathered}
\vect{A}^{(t)}_{ij}\ \big | \ \vect{F}^{(1)}, \ldots, \vect{F}^{(t)}  \stackrel{\text{ind}}{\sim} \text{Bernoulli}\big(\vect{Q}^{(t)}_{ij}\big)\\
\vect{Q}^{(t)}_{ij} = \lambda\vect{Q}^{(t-1)}_{ij} + (1-\lambda) \vect{P}^{(t)}_{ij}\qquad\text{and}\qquad
\vect{P}^{(t)}_{ij} = 	\begin{cases}\frac{1}{1+\exp\Big(-\vect{\beta}'\vect{F}^{(t)}_{ij} \Big)}  & \text{if \ $\vect{B}^{(t)}_{ij}=1$} \\\hfil 0 &  \text{otherwise}\end{cases}
\end{gathered}
\end{equation}
where $\lambda \in [0,1]$ controls the trade-off between the dependence on the past and the current feature vector for a link $(i,j)$. The model expresses our belief that given the features and link history, a connection at time $t$ is independent (but non-homogeneous) of all other links and that this probability evolves as an auto-regressive model with weights on probability of connection at time $t-1$ and a logit transformed probability induced by the features. We refer to the proposed model as \textit{Bernoulli Auto-Regressive (BAR)}. Note that the recursion of $\vect{Q}^{(t)}_{ij}$ in Model \eqref{eq:BAR} allows us to express the probability of connection at a given time as a weighted sum of past probabilities. This weighted sum along with a  non-linear dependence on the features provides flexibility in modeling arbitrary sequences with observation level features. This is more general than modeling the probabilities as a logit transform of lagged feature vectors. In the choice of $\vect{P}^{(t)}_{ij}$ we can also use different link functions to extend our model to sequences of discrete outcomes of $K$ categories where $K>2$ (e.g., by using the softmax function), providing that $\vect{P}^{(t)}_{ij}$ is valued between 0 and 1 so that $\vect{Q}^{(t)}_{ij}$ is a valid Bernoulli parameter.

For certain applications BAR may be under-parametrized. We can extend the model by having different coefficient vectors in the link probability for different sets of connections. These sets can be determined by using clustering on the features as a pre-processing step. More precisely we can have for $\vect{P}^{(t)}_{ij}$
\begin{align}
\vect{P}^{(t)}_{ij} &= \left\{
	\begin{array}{ll}
		 \frac{1}{1+\exp\left(-\vect{\beta}'_k\vect{F}^{(t)}_{ij} \right)}  & \mbox{if } \vect{B}^{(t)}_{ij}=1 \text{ and } C^{(t)}_{ij}=k \\
		 \hfil 0 & \text{otherwise}
	\end{array}
\right.
\label{eq:PtijCluster}
\end{align}
where $C^{(t)}_{ij}$ is the cluster membership for link $(i,j)$ at time $t$. We assume only one cluster for edge membership for the rest of the work.

\begin{figure}
\centering
\includegraphics[width=0.5\textwidth]{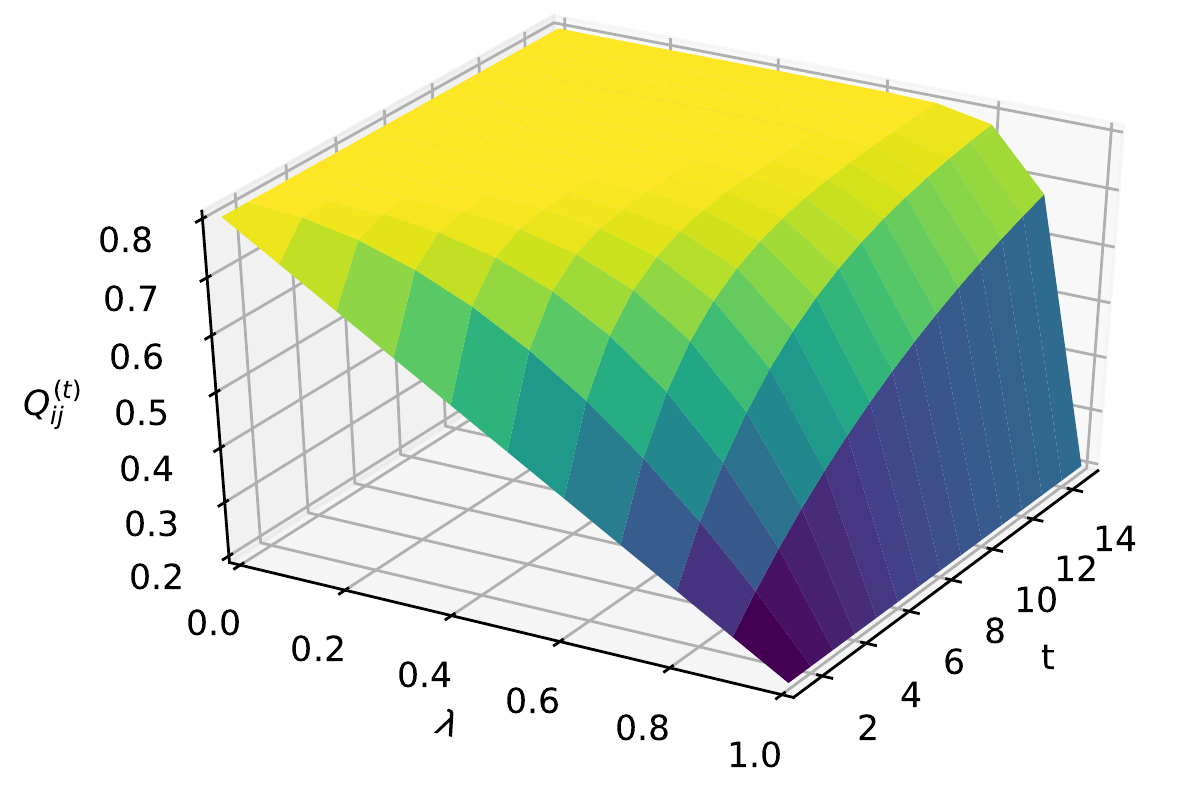}
\caption{Evolution of $\vect{Q}^{(t)}_{ij}$ over $t$ at various $\lambda$, under $\vect{Q}^{(0)}_{ij}=0.2$ and $\vect{P}^{(t)}_{ij}=0.8$}
\label{fig:Qt}
\end{figure}

The decaying contribution of historical features is more obvious once we express the recursion of $\vect{Q}^{(t)}_{ij}$ in Model \eqref{eq:BAR} equivalently as
\begin{equation}\label{eq:Qtrecursion}
\vect{Q}^{(t)}_{ij} = \lambda^t \vect{Q}^{(0)}_{ij} + \sum_{s=1}^t\lambda^{t-s}(1-\lambda)\vect{P}^{(s)}_{ij}.
\end{equation}
The longer the time has elapsed since time $s$, the smaller the weight ($\lambda^{t-s}(1-\lambda)$) is applied on the probability $\vect{P}^{(s)}_{ij}$ and its underlying features from time $s$. The diminishing contribution from features over elapsed time is a valid assumption in many applications (e.g., attribution model). The choice of $\lambda$ governs the decaying rate: larger $\lambda$ makes $\vect{Q}^{(t)}_{ij}$ more dependent on historical probabilities, whereas smaller $\lambda$ magnifies the dependence on current features. To illustrate the relation between $\vect{Q}^{(t)}_{ij}$ and $\lambda$, we consider a simplified scenario: we let $\vect{Q}^{(0)}_{ij}$ be 0.2 and $\vect{P}^{(t)}_{ij}$ be 0.8 for all $t\in \{1, \ldots, 15\}$. In Figure \ref{fig:Qt} we show how $\vect{Q}^{(t)}_{ij}$ evolves with $t$ at various $\lambda$ values. At extremes, $\vect{Q}^{(t)}_{ij}$ takes constant value of $\vect{Q}^{(0)}_{ij}$ (when $\lambda=1$) or $\vect{P}^{(t)}_{ij}$ (when $\lambda=0$). For $0<\lambda<1$, the larger $\lambda$ is, the more slowly the contribution of historical features to the $\vect{Q}^{(t)}_{ij}$ decays and the more slowly $\vect{Q}^{(t)}_{ij}$ deviates from its initial assignment (0.2) towards the probability derived from current features (0.8).

% Figure: Qt evolution

%%%%%%%%%%%%%%%%%%%%%%%%%%%%%%%%%
%                               BAR Model: Estimation                                       %
%%%%%%%%%%%%%%%%%%%%%%%%%%%%%%%%%
\subsection{Estimation}\label{sec:est}
Model estimation is performed by maximizing the log-likelihood function for Model \eqref{eq:BAR} since for each $(i,j)$ we can obtain a closed form expression for the probability of connection $\vect{Q}^{(t)}_{ij}$ at time $t$ using the recursive formula \eqref{eq:Qtrecursion}. However, as we mentioned before the main network can be very sparse and so we cannot hope to learn much from a small number of observations. To cast our network to a wider set of connections we add a regularization term to the log-likelihood and propose to minimize the following problem
\begin{align}\label{eq:regLik}
\underset{\vect{\beta} \in \Re^d}{\text{minimize}} \ \text{-Log-Likelihood} + \alpha \sum_{t=1}^T \left\| \left(\vect{Q}^{(t)} - \vect{B}^{(t)}\right) \vect{\Phi}^{(t)} \right\|^2_F
\end{align}
where $\vect{\Phi}^{(t)}=\vect{F}^{(t)}\otimes \vect{1}_N \in \Re^{N\times d}$ are aggregated features at sender-level, i.e., $\vect{\Phi}^{(t)}_{i\ell}=\sum_{\{j: \vect{B}^{(t)}_{ij}=1\}}\vect{F}^{(t)}_{ij\ell}$ encodes the $\ell$th feature aggregated over recipients of sender $i$.

The regularization term provides us a way to minimize deviation of first order statistics (for the features) between the network probabilities of interest and the observed auxiliary network. In another word, we want network aggregated features to be similar for the main and auxiliary sequences. The regularization and the probabilities $\vect{P}^{(t)}_{ij}$ are precisely what allow us to learn from auxiliary data, and Problem \eqref{eq:regLik} represents a trade-off between the task of link prediction and discovery. If we have a lot more observed data in the main network sequence and it is of interest to predict the prevalence of existing connections in the future we should give more weight to the log-likelihood term (choose a small $\alpha$) and if instead our goal is to discover newer connections we should focus on trying to minimize the regularizer term (with a larger $\alpha$). If there is only one network sequence with feature information, we simply take $\alpha=0$ to accommodate missing $\vect{B}^{(t)}$ and adjust $\vect{P}^{(t)}$'s dependence onto $\vect{A}^{(t)}$ (see Section \ref{sec:reddit} for an example). These adjustments would make our model generalizable to single network sequence with relevant features.

We propose a stochastic gradient descent (SGD) algorithm to solve Problem \eqref{eq:regLik}. Our SGD algorithm exploits data sparsity in the main and auxiliary networks and is scalable to networks with millions of nodes. We refer readers to Appendix \ref{appen:sgd} for details of the algorithm.

%%%%%%%%%%%%%%%%%%%%%%%%%%%%%%%%%
%                                   Simulation Study                                          %
%%%%%%%%%%%%%%%%%%%%%%%%%%%%%%%%%
\section{Simulation Study}\label{sec:simu}
We simulate data from the generative model \eqref{eq:BAR} and perform several experiments to gain more insight into its behavior and performance compared to competing models. Moreover, since the network sequences and edge features are highly likely to be statistically dependent, we employ settings that are indicative of real-world applications. To generate the first network in the auxiliary sequence we sample Erd\H{o}s-Renyi graphs with different combinations of $N$ and edge generation probabilities $p$. For subsequent time steps ($T=15$) we vary
\begin{align}
\vect{B}^{(t)}_{ij} &= \left\{
\begin{array}{ll}
\vect{B}^{(t-1)}_{ij}+1 \ \mbox{ w.p. } \ p_\text{add} \ \mbox{ if } \ \vect{B}^{(t-1)}_{ij}=0 \\
\vect{B}^{(t-1)}_{ij}-1 \ \mbox{ w.p. } \ p_\text{del} \ \mbox{ if } \ \vect{B}^{(t-1)}_{ij}=1. 
\end{array}
\right.
\label{eq:Btgen}
\end{align}
For every edge $\vect{B}^{(t)}_{ij}$ we sample the features $\vect{F}^{(t)}_{ij} \in \Re^{10}$ such that each element is $\text{Poisson}(\vect{\mu}_t)$ distributed, where $\vect{\mu}_t = \vect{\mu}_{t-1} + \vect{\epsilon}_t \sim \mathcal{N_{\text{10}}}(\vect{0},0.05\times\vect{I}_{10})$ is a simple random walk. The coefficient vector $\vect{\beta} \sim \text{Unif}[0,1]^{10}$ is normalized to have unit Euclidean norm. Finally, the main network sequence $\vect{A}^{(t)}_{ij}$ is generated by Bernoulli trials with probability $\vect{Q}^{(t)}_{ij}$ computed using the recursion in \eqref{eq:BAR}, where $\vect{Q}^{(0)}_{ij}$ is initialized as the maximum likelihood estimate of edge probability.

We compare the BAR model with two competing models. The first model is graph feature tracking (GFT) from \cite{richard2010} which incorporates a feature prediction and regularization procedure for estimating $\vect{A}^{(T+1)}$. In specific, \citet{richard2010} define $n\times k$ matrix of features $\vect{\Omega}$ that can capture the dynamics of $\vect{A}^{(t)}$, and project $\vect{A}^{(t)}$ with a linear feature map $\vect{G}^{(t)} = \vect{A}^{(t)}\vect{\Omega}$. With historical feature maps $\{\vect{G}^{(1)}, \ldots, \vect{G}^{(T)}\}$ \citet{richard2010} propose using ridge regression to estimate $\hat{\vect{G}}$ to substitute $\vect{A}^{(T+1)}\vect{\Omega}$. In the directed graph setting, we let the features $\vect{\Omega}=\vect{V}_{(:, 1:k)}\vect{\Sigma}^{-1}_{(:, 1:k)}$ where $\vect{V}_{(:, 1:k)}$ and $\sigma_{1:k}$ are $k$ leading singular vectors and singular values from the SVD of $\vect{A}^{(T)}=\vect{U}\vect{\Sigma}\vect{V}'$, respectively. We solve Problem \eqref{eq:GFT} to get a GFT estimator for $\vect{A}^{(T+1)}$:
%%%% GFT model
\begin{equation}\label{eq:GFT}
\min_{\vect{S}\in \Re^{n\times n}} \frac{1}{2}\left\| \vect{S} - \vect{A}^{(T)} \right\|_F^2 + \frac{1}{2}\nu \left\| \vect{S}\vect{\Omega} - \hat{\vect{G}} \right\|_F^2 + \tau \left\|\vect{S} \right\|_*
\end{equation}
where $\|\vect{S}\|_*$ is a nuclear norm for inducing low rankness in $\vect{S}$. Since the GFT estimator is not guaranteed to have its values between 0 and 1, we threshold any value of $\vect{S}$ outside $[0,1]$ to its closer boundary. In addition to GFT, we compare the BAR model to a logistic model with features being averaged historical features up to the current time.
%%%% logistic model
\begin{align}\label{logistic}
\vect{A}^{(t)}_{ij}\big | \ \vect{F}^{(1)}, \ldots, \vect{F}^{(t)} \stackrel{\text{ind}}{\sim} \text{Bernoulli} \left(\frac{1}{1+\exp\left( -\vect{\beta}'\vect{F}^{\text{avg}(t)}_{ij} \right)}  \right)
\end{align}
where averaged features across historical time $\vect{F}^{\text{avg}(t)}$ is defined as follows:
\begin{align}
\vect{F}^{\text{avg}(1)}_{ij} = \vect{F}^{(1)}_{ij}\quad \text{and for }t\ge2, \
\vect{F}^{\text{avg}(t)}_{ij} = \begin{cases} \vect{F}^{\text{avg}(t-1)}_{ij} \cdot (t-1)/t & \text{if }\vect{F}^{(t)}_{ij}\text{ does not exist}\\
\left(\vect{F}^{\text{avg}(t-1)}_{ij} \cdot (t-1) + \vect{F}^{(t)}_{ij}\right)/t & \text{if }\vect{F}^{(t)}_{ij}\text{ exists}.\end{cases}
\end{align}
 All three models under the comparison have access to historical data though they leverage the data differently.

%% Simulation table
\begin{table}[h]
\small
\centering
\caption{For each simulation scenario (Task, Parameter), Node Degree (No. Links/No. Nodes) for the main network in the test period and AUC ROC for each model (averaged over 10 repetitions).}
\tabcolsep=0.11cm
\begin{tabular}{ cccccc }
\toprule
\multirow{2}{*}{\begin{tabular}{c}Task\end{tabular}} & \multirow{2}{*}{\begin{tabular}{c}Parameter\end{tabular}} & \multirow{2}{*}{\begin{tabular}{c}Node\\Degree\end{tabular}} & \multicolumn{3}{c}{Models (AUC ROC)}\\
\cmidrule(lr){4-6}
& & & BAR & GFT & Logistic\\
\midrule
\multirow{3}{*}{\begin{tabular}{c}Standard\end{tabular}} 
& $N=n=10k, p=5\times 10^{-4}, p_{\text{add}}=10^{-3}p,p_{\text{del}}=10^{-2}p$ & 1.8 & \bf 0.972 & 0.936 & 0.505\\
& $N=n=10k, p=5\times 10^{-3}, p_{\text{add}}=10^{-3}p,p_{\text{del}}=10^{-2}p$ & 17.6& \bf 0.971 & 0.930 & 0.506\\
\midrule
\multirow{5}{*}{\begin{tabular}{c}Varying\\Network\\Dependence\end{tabular}} 
& $N=n=10k, p=5\times 10^{-3},p_{\text{add}}=10^{-3}p,p_{\text{del}}=0.0005$ & 17.6& \bf 0.970 & 0.929 & 0.508\\
& $N=n=10k, p=5\times 10^{-3}, p_{\text{add}}=10^{-3}p,p_{\text{del}}=0.005$ & 17.1& \bf 0.967 & 0.926 & 0.527\\
& $N=n=10k, p=5\times 10^{-3}, p_{\text{add}}=10^{-3}p,p_{\text{del}}=0.05$ & 13.3& \bf 0.941 & 0.893 & 0.648\\
& $N=n=10k, p=5\times 10^{-3}, p_{\text{add}}=10^{-3}p,p_{\text{del}}=0.2$ & 6.7 & \bf 0.901 & 0.799 & 0.725\\
& $N=n=10k, p=5\times 10^{-3}, p_{\text{add}}=10^{-3}p,p_{\text{del}}=0.5$ & 3.0 & \bf 0.855 & 0.700 & 0.670\\
\bottomrule
\end{tabular}
\label{tab:sim}
\end{table}

The comparisons are made over two tasks: a \textit{standard setting} where networks of different densities are considered, and a setting with \textit{varying network dependence} for which adjacent network variation is gradually increased. For each scenario, we fix $N$ and $n$ to 10 thousands and $p_{\text{add}}$ to $10^{-3}p$, but vary $p$ and $p_{\text{del}}$ over a range of values. We include all simulation settings in Table \ref{tab:sim}. For the task with varying network dependence, we gradually increase $p_{\text{del}}$ but fix all other parameters. Larger $p_{\text{del}}$ increases the probability of deleting an existing edge, injecting more variability into networks over time. By varying these parameters we highlight the advantages of our approach to a more traditional matrix completion framework (e.g., GFT from \cite{richard2010}) and also gain insight into the workings of our method. Under each parameter setting, we generate 10 data samples and fit each model onto the samples. We finely tune each model with training data from the first 14 periods and test on the last period. We report the averaged AUC ROC over 10 repetitions for each model in Table \ref{tab:sim}. On all tasks the BAR significantly outperforms the GFT and the logistic by achieving higher AUC ROC. When network becomes more varied and more sparse with increasing $p_{\text{del}}$, the challenge increases for all models; the BAR still constantly do better than the other two methods under this scenario.

%%%%%%%%%%%%%%%%%%%%%%%%%%%%%%%%%
%                                      Experiments                                            %
%%%%%%%%%%%%%%%%%%%%%%%%%%%%%%%%%
\section{Experiments}\label{sec:exp}
We exploit the performance of the BAR model on two real experiments. The first experiment uses product \textit{B} as an auxiliary network to predict and discover connections on product \textit{A} (the main network). The superior performance of the BAR model demonstrates the connections between the two underlying products. Moreover, the BAR model can fully leverage such connections to infer links on the product of interest (\textit{A}). Our second experiment focuses on sentiment classification with a Reddit hyperlink network. We construct temporal network sequences for hyperlinks and use the BAR model (along with the baselines) to classify sentiment for the Reddit posts. We show that the BAR model performs at least as well as the logistic baselines even when minimal time dependence is present in network sequence.

%%%% \textit{A} Application
\subsection{Link Prediction and Discovery with Real Products}\label{sec:A}
Products from large corporations often provide a connected ecosystem for users where the modality of connection is defined by a specific product or feature. For example, there could be a communication messaging and a social network. The different networks within such companies can have varying levels of connectivity. In our case product \textit{B}'s network is much bigger and denser than \textit{A}'s. Such an application is perfectly aligned with the framework we present here.

In the experiment we let the product \textit{A}'s graph correspond to the main network and the \textit{B}'s to the auxiliary network. The underlying intuition is that a network defining activity on product \textit{B} can be indicative of users who might be interested in the feature provided by \textit{A}. To prepare data, we extract directed links among users from \textit{A} and \textit{B} at weekly level. The network data covers a period of 32 weeks where the last week is reserved as a test set. We down sample the \textit{B}'s network for computational reasons and use 15 weekly edge-level features corresponding only to activity. Note that we are also unable to detail the exact data and size descriptions due to privacy concerns. Finally we tune the BAR model over a grid of $\alpha$ values between 0 and 1. For comparison, we use the same logistic model described in the simulation study in Section \ref{sec:simu} that accounts for time dependence on historical features. Since the GFT does not scale to large networks we are unable to make a comparison to it. We evaluate the model performance on the hold-out test set.

%%%%% Figure: Test Set Split
\begin{figure}
\centering
\begin{minipage}{0.31\textwidth}
\centering
%\textbf{Split of 1s in Test Set}
\includegraphics[width=0.9\textwidth]{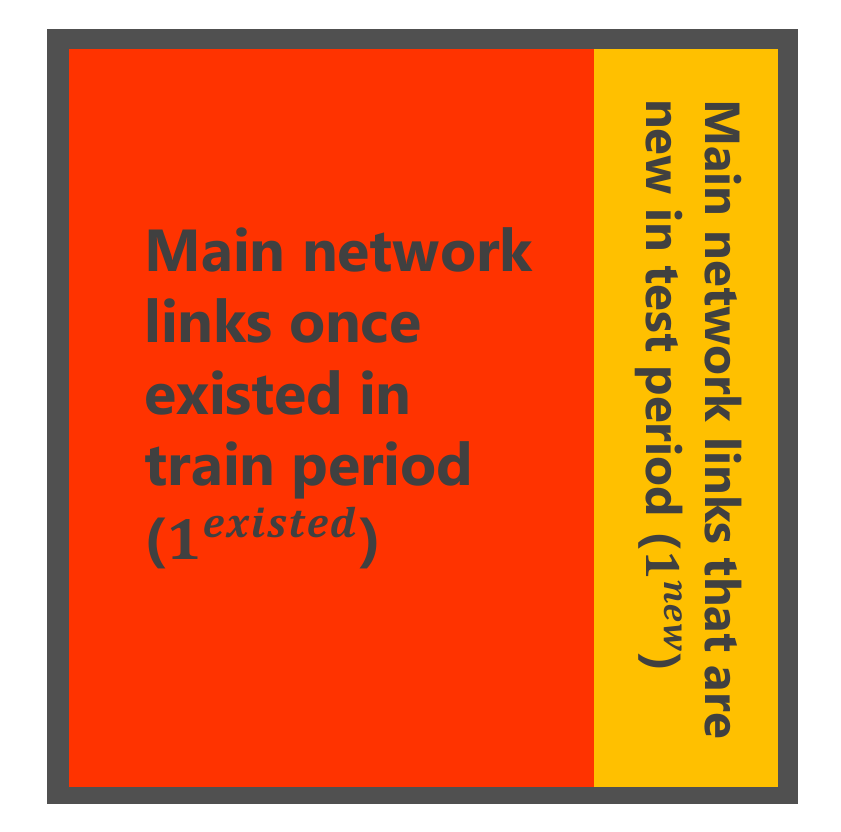}
\end{minipage}
\begin{minipage}{0.38\textwidth}
\centering
%\textbf{Split of 0s in Test Set}
\includegraphics[width=0.9\textwidth]{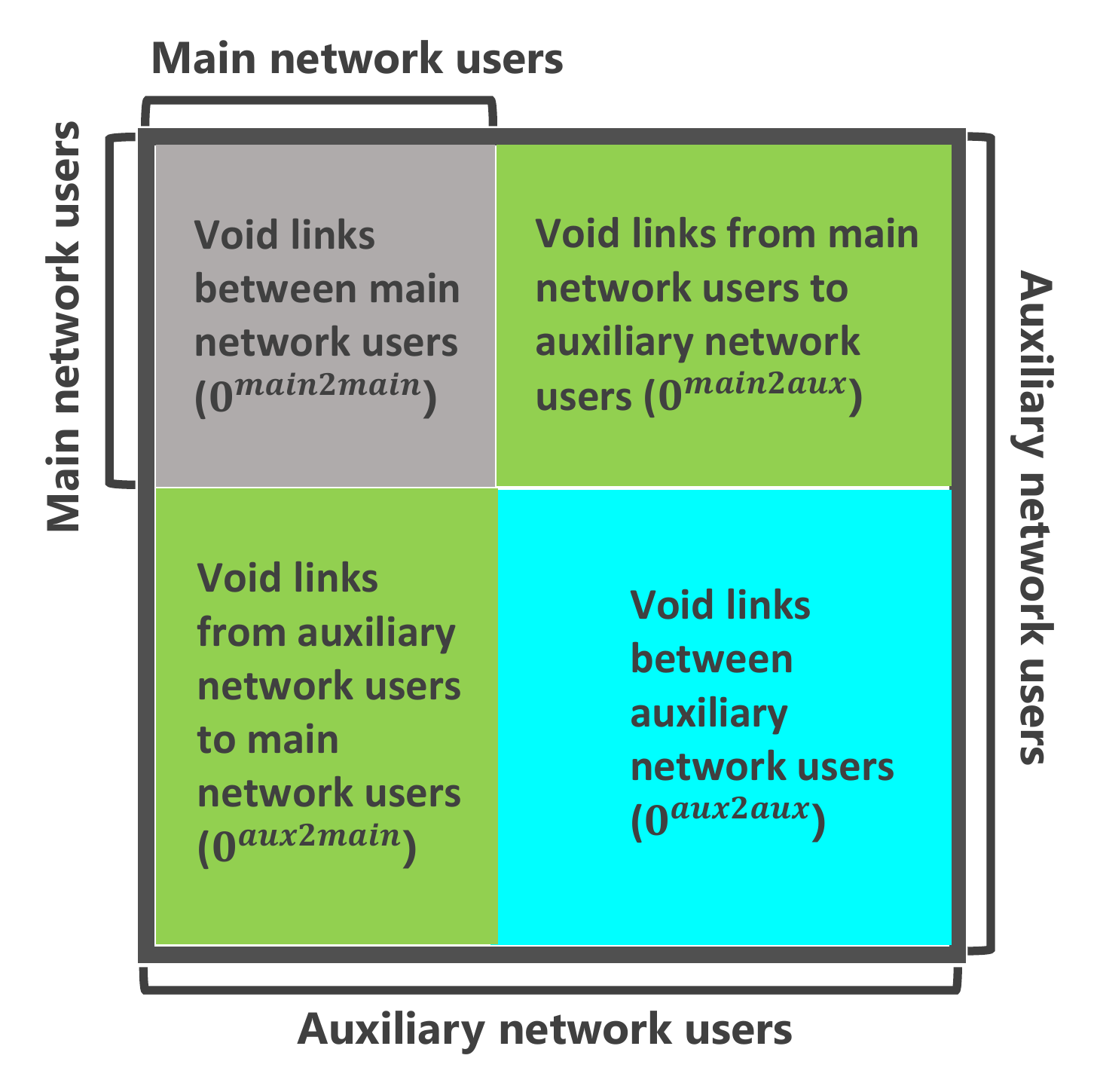}
\end{minipage}
\caption{Decomposition of the test set. (Left) Set of 1s (i.e., links) from the test period are partitioned into two subsets, depending on whether the link existed in train period ($1^{\text{existed}}$) or not ($1^{\text{new}}$). (Right) Set of 0s (i.e., non-link pairs) from the test period are partitioned into four subsets by user membership: between main network users ($0^{\text{main2main}}$), between auxiliary network users ($0^{\text{aux2aux}}$), from main network to auxiliary one and vice versa ($0^{\text{main2aux}}$ and $0^{\text{aux2main}}$).}
\label{fig:testset-split}
\end{figure}

We have two tasks at hand: \textit{link prediction} that aims at accurately predicting recurrent links, and \textit{link discovery} that targets on discovering new links that never existed before. We split the test set to properly evaluate the performance on the two tasks. The test set is made up of \textit{A}'s links from the last week (i.e., \textit{the set of ones}) and pair of users from \textit{A} or \textit{B} that did not connect on \textit{A} in the week (i.e., \textit{the set of zeros}). The test set is highly imbalanced: the set of zeros is larger than the set of ones by orders of magnitudes. From a practical standpoint, a predicted link on \textit{A} is more likely to hold if one of the two users already use \textit{A}. For link discovery, such an edge prediction can be directly translated into a recommendation setting. Hence, restricting to zero entries with exactly one user being an existing \textit{A} user would help improve prediction accuracy for \textit{any} model, and alleviate the imbalance in the test set. As is shown in the left panel of Figure \ref{fig:testset-split}, we partition the set of ones into links that existed in training (indexed as $1^{\text{existed}}$) and those that never existed before (indexed as $1^{\text{new}}$). The set $1^{\text{existed}}$ is suitable for gauging model performance on link prediction, whereas the set $1^{\text{new}}$ allows us to assess the link discovery task. In the right panel of Figure \ref{fig:testset-split}, we partition the set of zeros into four segments according to the memberships of the sender and receiver: $0^{\text{main2main}}, 0^{\text{aux2aux}}, 0^{\text{main2aux}}$ and $0^{\text{aux2main}}$. We refer readers to Figure \ref{fig:testset-split} for details of the partition. 

%% Table of AUC ROC for \textit{B} application
\begin{table}[h]
\small
\centering
\caption{Performance on link prediction and link discovery on \textit{A}'s network over test period.}
\tabcolsep=0.11cm
\begin{tabular}{ cllcc }
\toprule
\multirow{2}{*}{\begin{tabular}{c}Task\end{tabular}}
& \multicolumn{2}{c}{Composition of Test Set} & \multicolumn{2}{c}{Models (AUC ROC)}\\
\cmidrule(lr){2-3}
\cmidrule(lr){4-5}
& \multicolumn{1}{c}{Set of Ones} & \multicolumn{1}{c}{Set of Zeros} & BAR & Logistic\\
\midrule
Link Prediction & $1^{\text{existed}}$ & $0^{\text{main2aux}} \cup 0^{\text{aux2main}}$ & \bf 0.98 & 0.60\\
Link Discovery & $1^{\text{new}}$ & $0^{\text{main2aux}} \cup 0^{\text{aux2main}}$ & \bf 0.67 & 0.63\\
\bottomrule
\end{tabular}
\label{tab:perf-B}
\end{table}

%%%%% Figure: Histogram of Recall
\begin{figure}
\centering
\begin{minipage}{0.49\textwidth}
\centering
Recall among Top $N$ Predictions \\for Link Prediction\vspace{0.5em}
\includegraphics[width=1\textwidth]{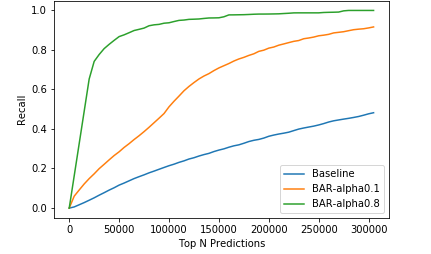}
\end{minipage}
\begin{minipage}{0.49\textwidth}
\centering
Recall among Top N Predictions \\for Link Discovery\vspace{0.5em}
\includegraphics[width=1\textwidth]{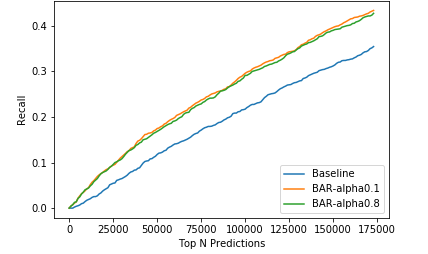}
\end{minipage}
\caption{Recall among top $N$ predictions for (Left) Link Prediction and (Right) Link Discovery. The top $N$ predictions are selected among $0^{\text{main2aux}}\cup 0^{\text{aux2main}}$ and the corresponding set of ones by task ($1^{\text{existed}}$ for link prediction and $1^{\text{new}}$ for link discovery).}
\label{fig:recall-topn}
\end{figure}

In general, link discovery is more challenging than link prediction since performance on the link discovery task is difficult to measure in the real setting specifically on offline data. This is because the generative mechanism of new links isn't known and our model assumption bases them on \textit{B} usage. A more reliable testing mechanism is perhaps randomized control trials. The contrast in the difficulty of the two taks is reflected in Table \ref{tab:perf-B}: the BAR model achieves almost perfect AUC ROC over link prediction while its corresponding metric for link discovery is much lower. Moreover, on both tasks the BAR model outperforms the logistic baseline. We also evaluate model performance with recall which is particularly suitable for evaluating link discovery. For each task, we compute recall from top $N$ predictions among the corresponding set of ones together with $0^{\text{main2aux}}\cup 0^{\text{aux2main}}$. As $N$ increases, we include more candidates which results in an improvement of recall for any model. In both panels of Figure \ref{fig:recall-topn}, the BAR model achieves significantly better recall than the baseline at every $N$. However, link discovery is more challenging which is reflected on the upper bound of the metric ($\sim$0.4) at large enough $N$ (see the right panel of Figure \ref{fig:recall-topn}). For link prediction (see the left panel of Figure \ref{fig:recall-topn}), the BAR model with $\alpha=0.8$ quickly converges to perfect recall. The very significant gap between the BAR curves and the logistic baseline curve from the left panel of Figure \ref{fig:recall-topn} points to the superiority of the BAR model.

%%%% Reddit Application
\subsection{Sentiment Classification on Reddit Hyperlink Network}\label{sec:reddit}
In the previous experiment we demonstrate significant improvement of the BAR model over a more standard approach in predicting and discovering links on networks with millions of nodes. Here we show that the proposed model is easily generalizable to \textit{any} dynamic binary classification task that spans over time. The Reddit hyperlink network\footnote{Reddit hyperlink network: \url{https://snap.stanford.edu/data/soc-RedditHyperlinks.html}} \cite{kumar2018} encodes the directed connections between two subreddit communities on \href{http://www.reddit.com/}{Reddit} from Jan 2014 to April 2017. The subreddit-to-subreddit hyperlink network is extracted from Reddit posts containing hyperlinks from one subreddit to another. The hyperlinks are time-stamped, directed and have attributes (a.k.a. features). Using crowd-sourcing and a text-based classifier, \citet{kumar2018} assigned a binary label to each hyperlink, for which label 1 indicates that the source expresses a positive sentiment towards the target and -1 if the source sentiment is negative towards the target.

%%%%% Figure: Desc stats (node and link) of Reddit hyperlink data
\begin{figure}
\centering
\begin{minipage}{0.43\textwidth}
\centering
\includegraphics[width=1\textwidth]{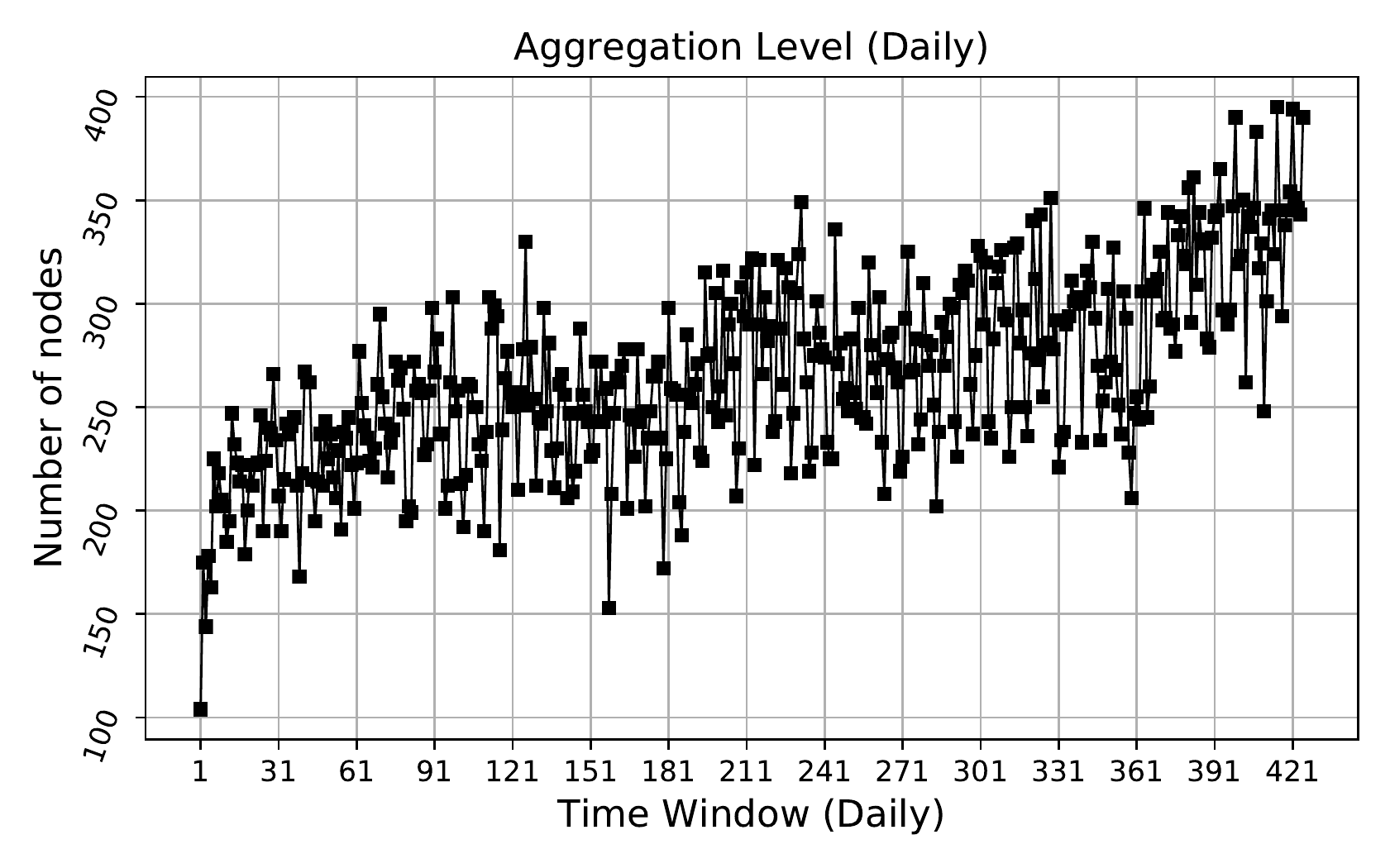}
\end{minipage}
\begin{minipage}{0.32\textwidth}
\centering
\includegraphics[width=1\textwidth]{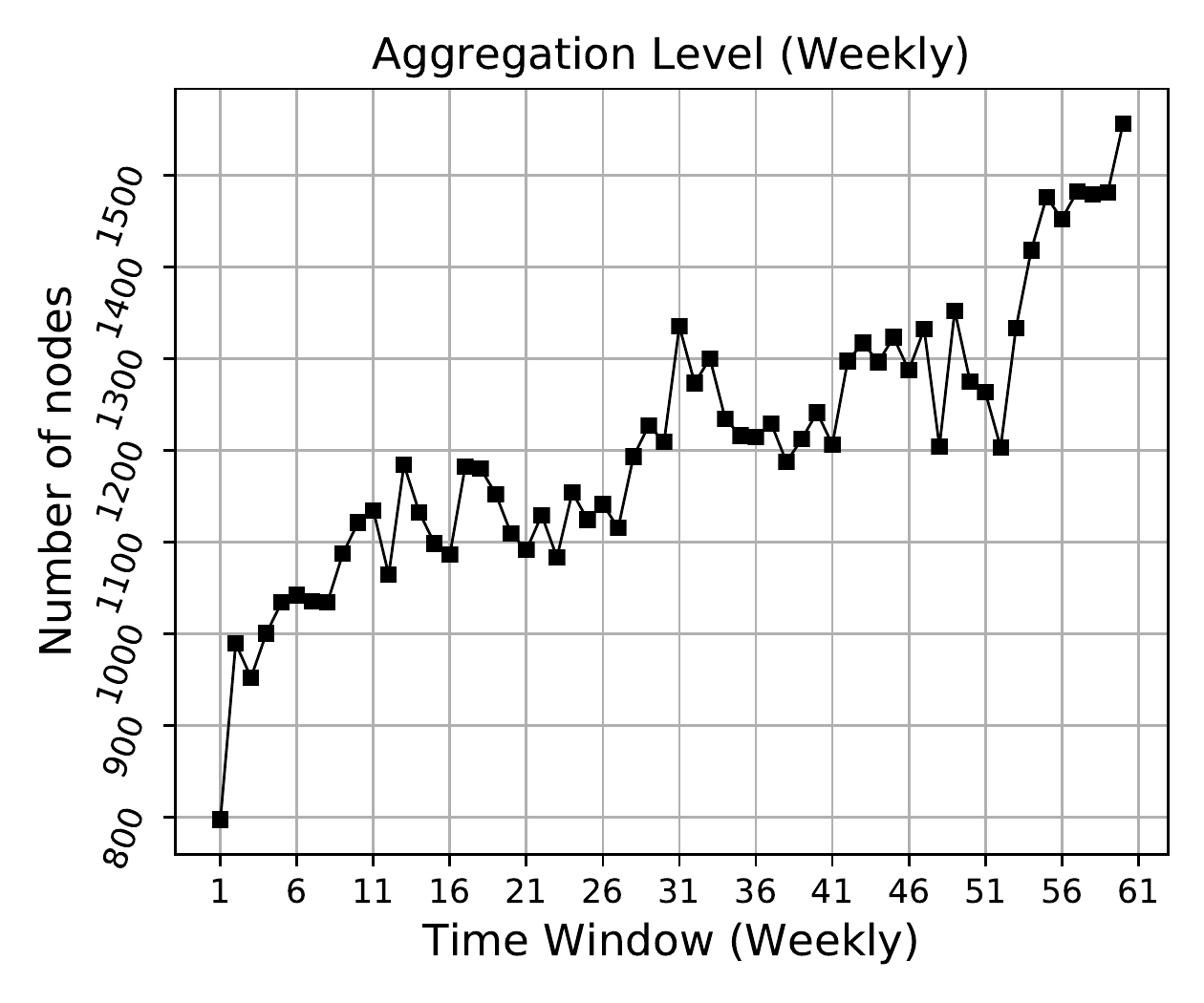}
\end{minipage}
\begin{minipage}{0.21\textwidth}
\centering
\includegraphics[width=1\textwidth]{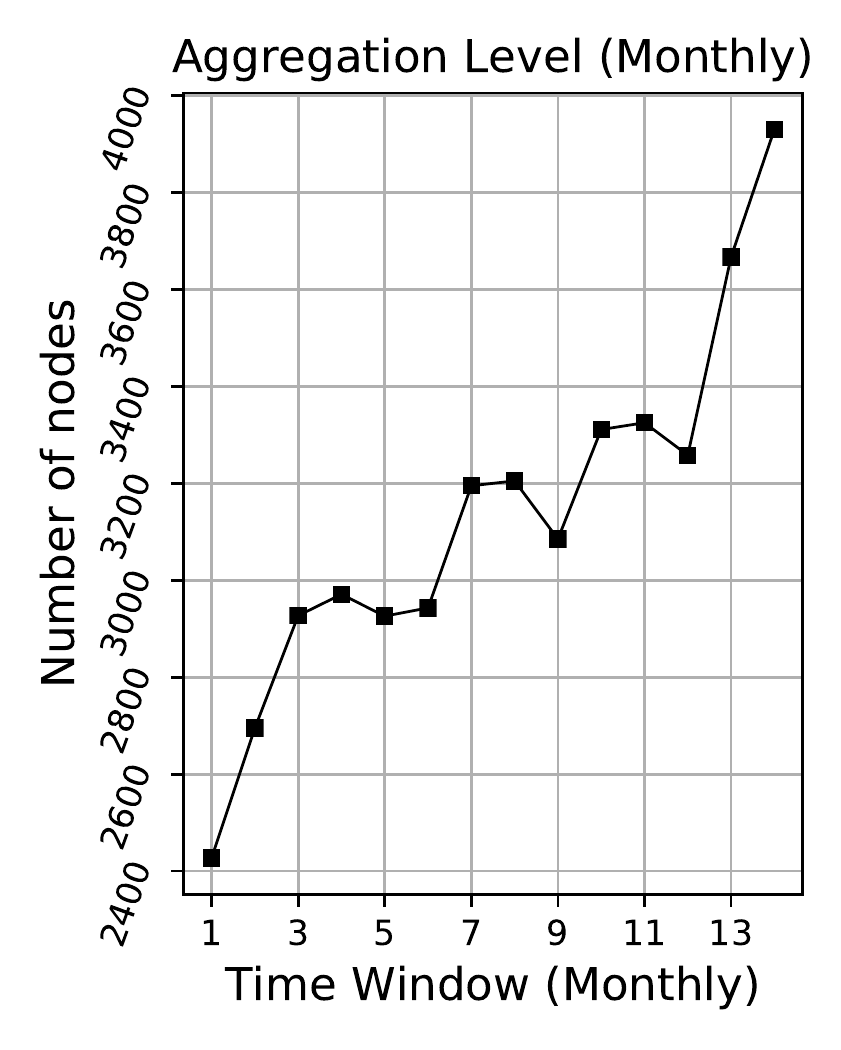}
\end{minipage}
\begin{minipage}{0.43\textwidth}
\centering
\includegraphics[width=1\textwidth]{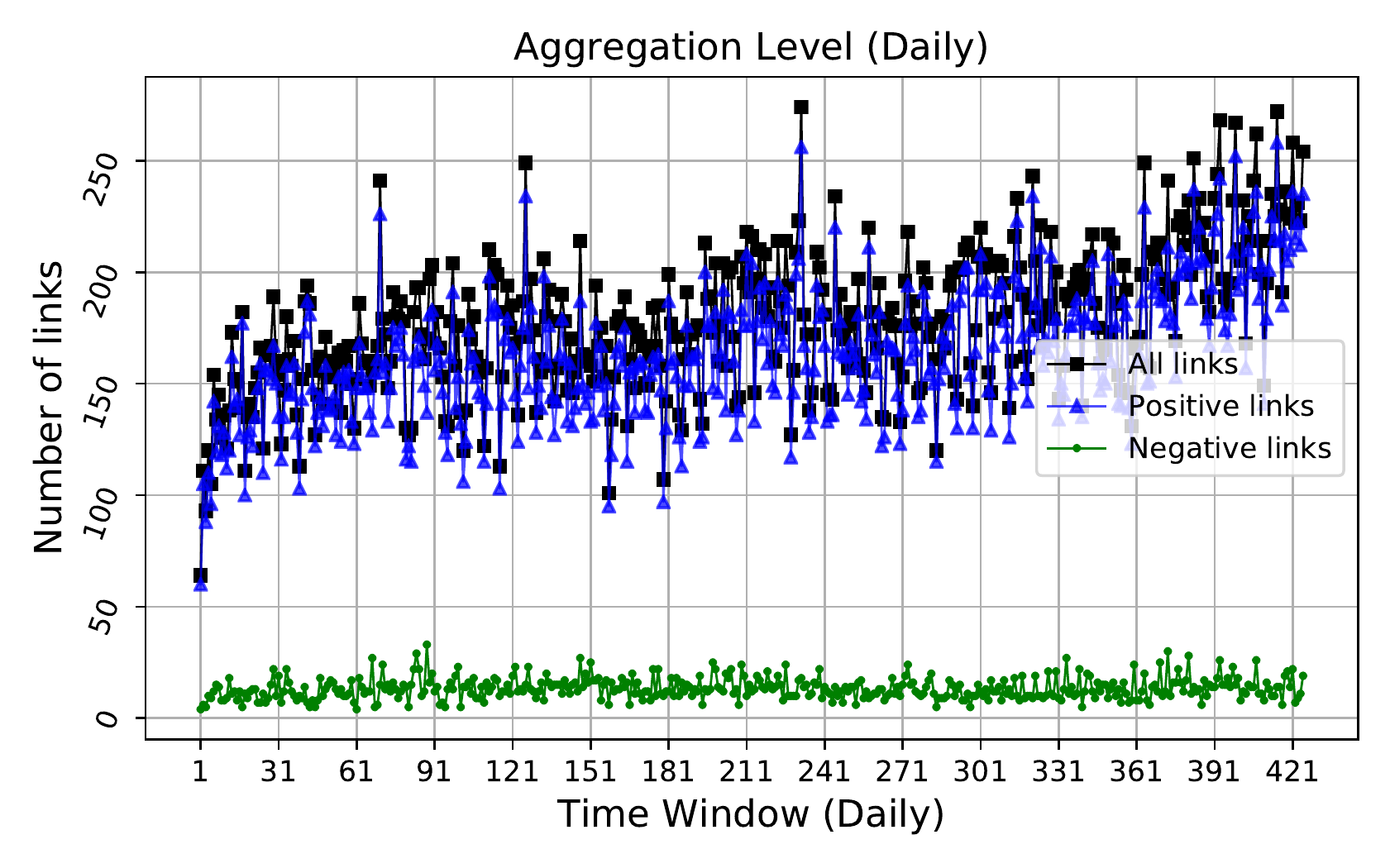}
\end{minipage}
\begin{minipage}{0.32\textwidth}
\centering
\includegraphics[width=1\textwidth]{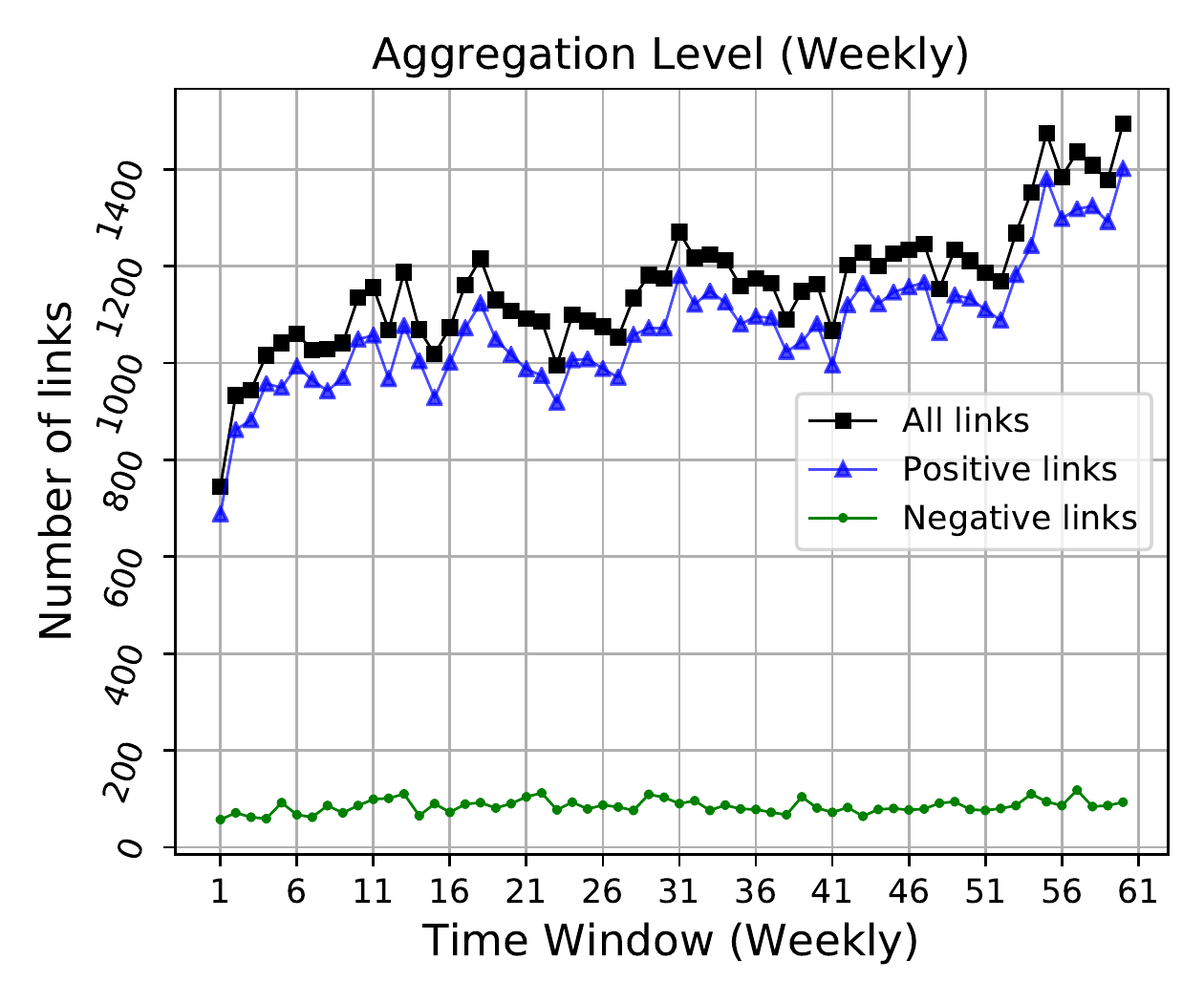}
\end{minipage}
\begin{minipage}{0.21\textwidth}
\centering
\includegraphics[width=1\textwidth]{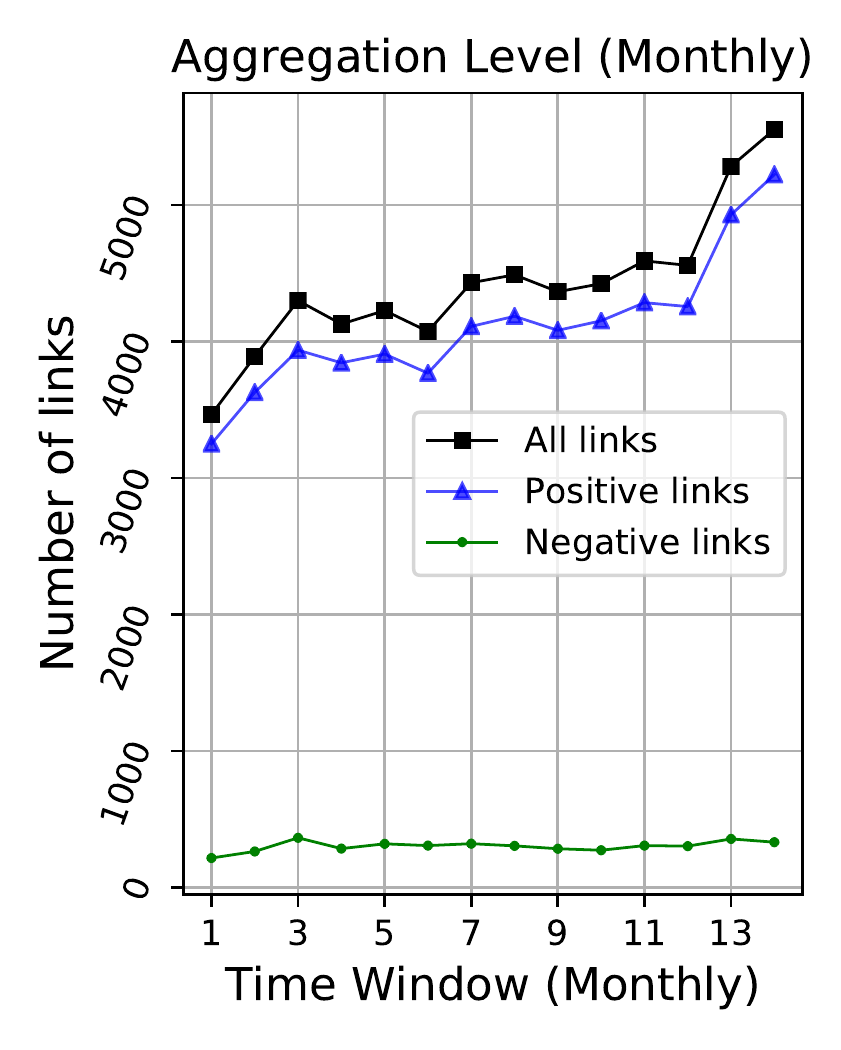}
\end{minipage}
\caption{Variation of the number of nodes (Top Row) and the number of links (Bottom Row) from Reddit hyperlink network over time at different aggregation levels: (Left) Daily, (Middle) Weekly and (Right) Monthly.}
\label{fig:reddit-desc}
\end{figure}

Our goal for the experiment is to classify sentiment for the hyperlinks with the BAR model and competing ones. The BAR model is designed for discrete times; however, the time stamps in the hyperlink network are continuous. To discretize the time stamps, we aggregate the hyperlinks at various granularities: daily, weekly and monthly. At each granularity we average the feature vectors corresponding to the same pair of (source subreddit, target subreddit) within each time window. Among the 86 features (see \cite{kumar2018} for details) we omit 11 of them that are count features because they are likely to be dominated by longer posts after the averaging. For sentiment labels within each time window, we take the majority vote as the label for (source subreddit, target subreddit). We experiment on the first 14 months of data (or roughly equivalently, the first 60 weeks or 425 days) from Jan 2014 to Feb 2015. Figure \ref{fig:reddit-desc} shows how the number of nodes and the number of links vary over time at different time granularities. At all three levels (daily, weekly and monthly) the number of nodes and links have an overall increasing trend over time. From the bottom row of Figure \ref{fig:reddit-desc} we observe the highly imbalance of the sentiment labels: there are much more positive hyperlinks than negative ones at any of the time granularities. As we equally value positive and negative hyperlinks, we use AUC ROC to evaluate the model performance over test period which is reserved for the last two months (or roughly equivalently, data from the last 8 weeks or 60 days).

%% Table of AUC ROC for Reddit application
\begin{table}[h]
\small
\centering
\caption{Performance of different models on Reddit hyperlink sentiment classification over test period. Between the two statistics, Node Degree (No. Links/No. Nodes) reflects the density of the network and \% Recurrent Links (Percentage of links that appear more than once over the time) reflects the time dependency structure.}
\tabcolsep=0.11cm
\begin{tabular}{ cccccccc }
\toprule
\multirow{2}{*}{\begin{tabular}{c}Aggregation\\Level\end{tabular}} 
& \multicolumn{2}{c}{Network Statistics} & \multicolumn{2}{c}{Train/Test Periods} & \multicolumn{3}{c}{Models (AUC ROC)} \\ 
\cmidrule(lr){2-3}
\cmidrule(lr){4-5}
\cmidrule(lr){6-8}
& Node Degree & \% Recurrent Links & $T_{\text{train}}$ & $T_{\text{test}}$ & BAR & Logistic (Raw) & Logistic\\
\midrule
Daily & 5.29 & 21.60\% & 365 & 60 & \bf 0.761 & 0.756 & 0.744 \\
Weekly & 4.96 & 20.99\% & 52 & 8 & \bf 0.760 & 0.755 & 0.743 \\
Monthly & 4.34 & 19.29\% & 12 & 2 & \bf 0.756 & 0.748 & 0.745 \\
\bottomrule
\end{tabular}
\label{tab:perf-reddit}
\end{table}

We modify the BAR model to adapt to the single network scenario. Let $\vect{A}^{(t)}_{ij}\in \{-1, 1\}$ represent the sentiment from the $i$th subreddit to the $j$th subreddit at time window $t$. We require the edge-level features $\vect{F}^{(t)}_{ij}$ to be from the network and update the dependency of $\vect{P}^{(t)}_{ij}$ onto $\vect{A}^{(t)}_{ij}$
\begin{equation}\label{eq:BAR-singlenetwork}
\vect{P}^{(t)}_{ij} = \begin{cases}\frac{1}{1+\exp\Big(-\vect{\beta}'\vect{F}^{(t)}_{ij} \Big)} & \text{if \ $\vect{A}^{(t)}_{ij}\in \{1, -1\}$} \\\hfil 0 & \text{otherwise}.
\end{cases}
\end{equation}
With single network we turn off the link discovery option by setting $\alpha$ to zero in Problem \eqref{eq:regLik}. For comparison we consider two variations of the logistic model. The first variation keeps using averaged historical features for classifying sentiment at the current time window, as is used Section \ref{sec:simu} and Section \ref{sec:A}, which we label as \textit{Logistic} in Table \ref{tab:perf-reddit}. The second variation of the logistic model, which we label with \textit{Logistic (Raw)} in Table \ref{tab:perf-reddit}, simply use the features from the current time window for classifying sentiment. Table \ref{tab:perf-reddit} has a comparison of the performance on corresponding test period across varying aggregation level. As we aggregate over longer time window (from daily to monthly), the network becomes sparser with node degree continuously decreasing. Comparing the two variations of the logistic model, it performs better without averaging historical features which indicates there is limited time dependence in the sentiment classification task. Indeed, our approach performs the best with daily aggregation when there is more time dependence, as is evidenced by the highest percentage of recurrent links. At all aggregation levels, the BAR model outperforms the logistic ones, showing its strength over binary classification task even with limited time dependence. As we mentioned before the competing GFT approach does not scale to large datasets. Moreover, the sentiment classification problem is not directly solvable with GFT's matrix completion framework in \eqref{eq:GFT}. Thus, we left the GFT approach out in this comparison.

%%%%%%%%%%%%%%%%%%%%%%%%%%%%%%%%%
%                                      Conclusions                                            %
%%%%%%%%%%%%%%%%%%%%%%%%%%%%%%%%%
\section{Conclusions}\label{sec:conclude}
We proposed a generalized auto-regressive model for link prediction and discovery with network sequences. The proposed BAR model can exploit signals from the auxiliary network to infer connections on the main network. We also develop an optimization framework for estimating the model, and a stochastic gradient descent algorithm that is scalable to networks with millions of nodes. To demonstrate the superior performance of our model, we apply the BAR model along with baselines on both simulated data and real data, and show significant improvement in BAR's performance over baselines. Our experiment on the product \textit{A}'s network shows practical use of the BAR model at scale, while the Reddit hyperlink experiment validates the generalizability of BAR. For future line of research, an extension of the BAR model to continuous time would be useful for many real-world applications. We also plan to investigate the theoretical properties of our model.

%%%%%%%%%%%%%%%%%%%%%%%%%%%%%%%%%
%                      Acknowledgements                          %
%%%%%%%%%%%%%%%%%%%%%%%%%%%%%%%%%
\section*{Acknowledgements}
The authors thank Sushama Murthy and Richard Johnston for supporting the work.

%%%%%%%%%%%%%%%%%%%%%%%%%%%%%%%%%
%                      Appendices                          %
%%%%%%%%%%%%%%%%%%%%%%%%%%%%%%%%%
\appendix
\section*{Appendix}
\addcontentsline{toc}{section}{Appendix}
\renewcommand{\thesubsection}{\Alph{subsection}}

%%%% Appendix: SGD
\subsection{Algorithm for BAR Model Optimization}\label{appen:sgd}
We derive the log-likelihood part and the regularizer part of Problem \eqref{eq:regLik} as follows.
\begin{align}
\text{Log-Likelihood} &=\sum_{t=1}^T\left(\sum_{\left\{(i,j): \vect{A}^{(t)}_{ij}=1\right\}} \log\left(\vect{Q}^{(t)}_{ij} \right)    + \sum_{\left\{(i,j): \vect{A}^{(t)}_{ij}=0\right\}} \log\left(1-\vect{Q}^{(t)}_{ij} \right)        \right)\label{part:loglik}\\
\text{Regularizer} &=\sum_{t=1}^T\sum_{i=1}^N\sum_{\ell=1}^d \left( \sum_{j=1}^N \left(\vect{Q}^{(t)}_{ij} - \vect{B}^{(t)}_{ij}     \right)\cdot\vect{\Phi}^{(t)}_{j\ell}    \right)^2\label{part:regularizer}
\end{align}
For the log-likelihood part \eqref{part:loglik}, we represent the summands using $f(\vect{\beta},t,i,j)$ and $g(\vect{\beta},t,i,j)$ defined as follows:
\begin{itemize}
\item $f(\vect{\beta}, t, i, j) :=\log\left(\vect{Q}^{(t)}_{ij}\right) = \log\left( \lambda^t \vect{Q}^{(0)}_{ij} + \sum_{s=1}^t\lambda^{t-s}(1-\lambda)\vect{P}^{(s)}_{ij} \right)$ when $\vect{A}^{(t)}_{ij}=1$;
\item $g(\vect{\beta}, t, i, j):=\log\left( 1-\vect{Q}^{(t)}_{ij}\right) = \log\left( 1-\lambda^t \vect{Q}^{(0)}_{ij} - \sum_{s=1}^t\lambda^{t-s}(1-\lambda)\vect{P}^{(s)}_{ij} \right)$ when $\vect{A}^{(t)}_{ij}=0$.
\end{itemize}
The gradients of $f(\vect{\beta}, t, i, j)$ and $g(\vect{\beta}, t, i, j)$ can be derived by the Chain Rule:
\begin{itemize}
\item $\nabla_{\vect{\beta}}f(\vect{\beta},t,i,j) =\frac{1}{\lambda^t \vect{Q}^{(0)}_{ij} + \sum_{s=1}^t\lambda^{t-s}(1-\lambda)\vect{P}^{(s)}_{ij}} \cdot \sum_{s=1}^t \lambda^{t-s}(1-\lambda)\nabla_{\vect{\beta}}\vect{P}^{(s)}_{ij}$;
\item $\nabla_{\vect{\beta}}g(\vect{\beta},t,i,j) =\frac{-1}{1-\lambda^t \vect{Q}^{(0)}_{ij} - \sum_{s=1}^t\lambda^{t-s}(1-\lambda)\vect{P}^{(s)}_{ij}} \cdot \sum_{s=1}^t \lambda^{t-s}(1-\lambda)\nabla_{\vect{\beta}}\vect{P}^{(s)}_{ij}$;
\end{itemize}
where $\nabla_{\vect{\beta}}\vect{P}^{(s)}_{ij}=\vect{P}^{(s)}_{ij}\left(1-\vect{P}^{(s)}_{ij}\right)\vect{F}^{(s)}_{ij}$ if $\vect{B}^{(s)}_{ij}=1$ and 0 otherwise. The gradient of the log-likelihood is 
\begin{equation*}
\nabla_{\vect{\beta}}\text{Log-Likelihood} = \sum_{t=1}^T \left(\sum_{\left\{(i,j): \vect{A}^{(t)}_{ij}=1\right\}} \nabla_{\vect{\beta}} f(\vect{\beta}, t,i,j) + \sum_{\left\{(i,j): \vect{A}^{(t)}_{ij}=0\right\}} \nabla_{\vect{\beta}} g(\vect{\beta},t,i,j)    \right).
\end{equation*}
For the regularizer part \eqref{part:regularizer}, we represent the summand using $h(\vect{\beta}, t,i,\ell) = \left(\sum_{j=1}^N \left( \vect{Q}^{(t)}_{ij} - \vect{B}^{(t)}_{ij} \right)\cdot \vect{\Phi}^{(t)}_{j\ell}  \right)^2$ of which the gradient can be derived by the Chain Rule as
\begin{equation*}
\nabla_{\vect{\beta}}h(\vect{\beta},t,i,\ell)=2\left(\sum_{j=1}^N\left( \vect{Q}^{(t)}_{ij}-\vect{B}^{(t)}_{ij} \right) \cdot \vect{\Phi}^{(t)}_{j\ell} \right)  \cdot \left(\sum_{j=1}^N \vect{\Phi}^{(t)}_{j\ell} \sum_{s=1}^t\lambda^{t-s}(1-\lambda)\nabla_{\vect{\beta}}\vect{P}^{(s)}_{ij} \right).
\end{equation*}
The gradient of the regularizer is 
\begin{equation*}
\nabla_{\vect{\beta}} \text{Regularizer} = \sum_{t=1}^T\sum_{i=1}^N\sum_{\ell=1}^d \nabla_{\vect{\beta}}h(\vect{\beta},t,i,\ell)
\end{equation*}
A gradient descent update at the $k$th iteration with step size $\eta$ has the form 
\begin{equation*}
\vect{\beta}^{k} \leftarrow \vect{\beta}^{k-1} - \eta\left(-\nabla_{\vect{\beta}}\text{Log-Likelihood} + \alpha\cdot\nabla_{\vect{\beta}}\text{Regularizer}  \right).
\end{equation*}
We propose a computationally feasible substitute to the gradient descent. At the $k$th iteration, we randomly sample a time stamp $t$, a sender node $i_{\text{lk}}$ for the log-likelihood gradient, and a sender node $i_{\text{reg}}$ for the regularizer gradient. The Stochastic Gradient Descent (SGD) algorithm is detailed in Algorithm \ref{alg:sgd}. To efficiently compute $\texttt{grad}_{\texttt{lk}}$ and $\texttt{grad}_{\texttt{reg}}$ from Lines 8-9 of the algorithm, we leverage the sparsity in the auxiliary sequence $\vect{B}^{(s)}$: for a given $i$ there are limited $j$s with $\vect{B}^{(s)}_{ij} = 1$. Hence, only a small number of non-zero $\nabla_{\vect{\beta}}\vect{P}^{(s)}_{ij}$ needs to be accounted for in computing the gradients of $f(\cdot), g(\cdot)$ and $h(\cdot)$.

%%%% SGD Algorithm
\begin{algorithm}
\caption{\em Stochastic Gradient Descent Algorithm for Solving Problem \eqref{eq:regLik}}
\begin{algorithmic}[1]
\Input $\vect{A}^{(t)}, \vect{B}^{(t)}, \vect{F}^{(t)}, \vect{\Phi}^{(t)}, \lambda, \alpha$ and $\eta$
\State Initialize $\vect{\beta^0}$
\State $k\leftarrow 0$
\While{$\vect{\beta}^k$ not converge}
\State $k\leftarrow k+1$
\State Sample $t$ from $\{1,\dots, T\}$
\State Sample $i_{\text{lk}}$ from $\left\{i:\vect{A}^{(t)}_{ij} = 1\text{ for some }j\right\}$
\State Sample $i_{\text{reg}}$ from $\left\{i:\vect{B}^{(t)}_{ij} = 1\text{ for some }j\right\}$
\State $\texttt{grad}_{\texttt{lk}}\leftarrow\sum_{\left\{j:\vect{A}^{(t)}_{i_{\text{lk}}j}=1 \right\}}\nabla_{\vect{\beta}}f\left(\vect{\beta}^{k-1}, t, i_{\text{lk}}, j\right)
+\sum_{\left\{j:\vect{A}^{(t)}_{i_{\text{lk}}j}=0 \right\}}\nabla_{\vect{\beta}}g\left(\vect{\beta}^{k-1}, t, i_{\text{lk}}, j\right)$
\State $\texttt{grad}_{\texttt{reg}}\leftarrow\sum_{\ell=1}^d \nabla_{\vect{\beta}}h\left(\vect{\beta}^{k-1}, t, i_{\text{reg}}, \ell\right)$
\State $\vect{\beta}^{k}\leftarrow \vect{\beta}^{k-1} - \eta \left(-\texttt{grad}_{\texttt{lk}} + \alpha\cdot \texttt{grad}_{\texttt{reg}} \right)$
\EndWhile
\Output $\vect{\beta}^k$
\end{algorithmic}
\label{alg:sgd}
\end{algorithm}

%% Make Appendices automatically
\bibliography{ref}
\bibliographystyle{apalike}
\end{document}